\documentclass[twocolumn]{bmcart}%
\usepackage{graphicx}
\usepackage{makecell}
\usepackage{tabularx}
\usepackage[american]{babel}

\usepackage{amsthm,amsmath,amssymb}

\usepackage{url}

\usepackage[utf8]{inputenc} %
\startlocaldefs
\endlocaldefs

\begin{document}

\begin{frontmatter}

\begin{fmbox}
\dochead{Research Articles}

\title{TasselNet: Counting maize tassels in the wild via local counts regression network
}

\author[
   addressref={aff1},                   %
   email={poppinace@hust.edu.cn}        %
]{\inits{HL}\fnm{Hao} \snm{Lu}}
\author[
   addressref={aff1},
   corref={aff1},                       %
   email={zgcao@hust.edu.cn}
]{\inits{ZC}\fnm{Zhiguo} \snm{Cao}}
\author[
addressref={aff1},
email={Yang\_Xiao@hust.edu.cn}
]{\inits{ZC}\fnm{Yang} \snm{Xiao}}
\author[
addressref={aff2},
email={bohan.zhuang@adelaide.edu.au}
]{\inits{CS}\fnm{Bohan} \snm{Zhuang}}
\author[
addressref={aff2},
email={chunhua.shen@adelaide.edu.au}
]{\inits{CS}\fnm{Chunhua} \snm{Shen}}

\address[id=aff1]{%
  \orgname{National Key Laboratory of Science and Technology on Multi-Spectral Information Processing, School of Automation, Huazhong University of Science and Technology}, %
  \city{Wuhan},                               %
  \postcode{430074},                          %
  \cny{China}                              %
}
\address[id=aff2]{%
  \orgname{School of Computer Science, The University of Adelaide},
  \city{Adelaide},
  \postcode{SA 5005},
  \cny{Australia}
}

\begin{artnotes}
{Part of this work was done when
the first author was visiting The University of Adelaide, Australia.}
\end{artnotes}

\begin{abstractbox}

\begin{abstract} %

\parttitle{Background} %
Accurately counting maize tassels is important for monitoring the growth status of maize plants. This tedious task, however, is still mainly done by manual efforts. In the context of modern plant phenotyping, automating this task is required to meet the need of large-scale analysis of genotype and phenotype. In recent years, computer vision technologies have experienced a significant breakthrough due to the emergence of large-scale datasets and increased computational resources. Naturally image-based approaches have also received much attention in plant-related studies. Yet a fact is that most image-based systems for plant phenotyping are deployed under controlled laboratory environment. When transferring the application scenario to unconstrained in-field conditions, intrinsic and extrinsic variations in the wild pose great challenges for accurate counting of maize tassels, which goes beyond the ability of conventional image processing techniques. This calls for further robust computer vision approaches to address in-field variations.

\parttitle{Results} %
This paper studies the in-field counting problem of maize tassels. To our knowledge, this is the first time that a plant-related counting problem is considered using computer vision technologies under unconstrained field-based environment. With 361 field images collected in four experimental fields across China between 2010 and 2015 and corresponding manually-labelled dotted annotations, a novel \textit{Maize Tassels Counting (MTC)} dataset is created and will be released with this paper. To alleviate the in-field challenges, a deep convolutional neural network-based approach termed \textit{Tasselnet} is proposed. Tasselnet can achieve good adaptability to in-field variations via modelling the local visual characteristics of field images and regressing the local counts of maize tassels. Extensive results on the MTC dataset demonstrate that Tasselnet outperforms other state-of-the-art approaches by large margins and achieves the overall best counting performance, with a mean absolute error of $6.6$ and a mean squared error of $9.6$ averaged over $8$ test sequences.

\parttitle{Conclusions}
Tasselnet can achieve robust in-field counting of maize tassels with a relatively high degree of accuracy. Our experimental evaluations also suggest several good practices for practitioners working on maize-tassel-like counting problems. It is worth noting that, though the counting errors have been greatly reduced by the Tasselnet, in-field counting of maize tassels remains an open and unsolved problem.

\end{abstract}


\end{abstractbox}
\end{fmbox}%

\end{frontmatter}

\section*{Background}
We consider the problem of counting maize tassels from images captured in the field using computer vision. Maize tassels are the male flowers of maize plants. The emergence of tassels indicates the arrival of the reproductive stage. During this stage, the total tassel number is an important cue to monitor the growth status of maize plants. It is closely related to the growth stage~\cite{ye2013image}, flowering time~\cite{guo2015automated}, and yield potential~\cite{lu2015fine}. In practice, counting maize tassels still mainly depends on human efforts, which is inefficient and fallible. Such a tedious task should be replaced by machines in modern plant phenotyping.

\begin{figure*}[h!]
	\centering
	\includegraphics[width=0.9\textwidth]{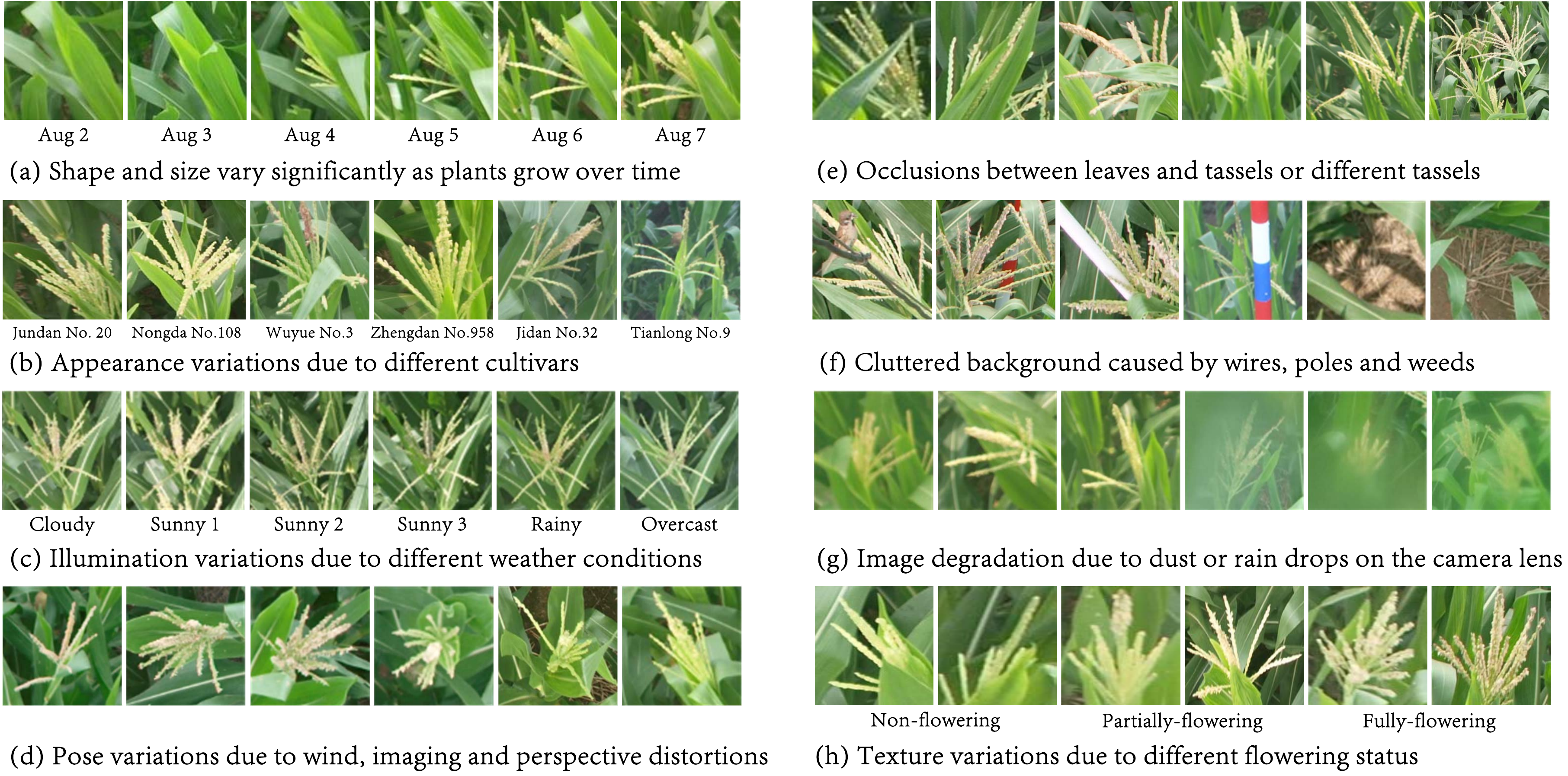}
	\caption{Intrinsic and extrinsic variations in the maize field. These variations pose significant challenges for in-field counting of maize tassels.}
	\label{fig:challenges}
\end{figure*}

To meet the need of large-scale and high-throughput analysis in plant phenotyping, image-based techniques provide a feasible, low-end, and efficient solution and have thus received much attention recently~\cite{guo2015automated,lu2015fine,yang2014combining,gage2017tips}. However, most practitioners and researchers still conduct experiments under controlled artificial environment. Although indoor experiments do simplify the process of image processing and advance our knowledge regarding the link between genotype and phenotype, ultimately plant phenotyping must be transferred to real-world scenarios, such as in the field or greenhouse~\cite{fiorani2013future}. Unfortunately, intrinsic and extrinsic variations in the wild field render the understanding and processing of field-based images a challenging task. Such challenges become more serious in the problem of in-field counting of maize tassels. As shown in Fig.~\ref{fig:challenges}, these challenges can largely boil down to the variations in the field-based environment:
\begin{itemize}
	\item Maize tassels emerge suddenly and vary significantly in shape and size as plants grow over time;
	\item Different cultivars of maize plants exhibit different appearance variations, such as colour and texture;
	\item Illumination changes dramatically due to different weather conditions, especially during the sunny day;
	\item The wind, imaging angle and perspective distortions cause various posture variations;
	\item Occlusions occur frequently, which renders the difficulty for counting even for a human expert;
	\item The cluttered background make visual patterns of maize tassels diverse and misleading;
	\item The quality of images degrades because of the dust or rain drops on the camera lens;
	\item Textural patterns also change essentially due to different flowering status.
\end{itemize}
It is worth noting that these challenges are not only specific to maize tassels but also applicable to a wide species of plants. It is inevitable to face these in-field variations before deploying plant phenotyping systems in the wild.

Though efforts have been made to tackle above problems and have achieved a moderate degree of success, the precision of the state-of-the-art tassel detection method is still below 50\%~\cite{lu2015fine}. This may be largely due to the inherent limitation of the \emph{non-maximum suppression} mechanism within object detection~\cite{felzenszwalb2010dpm}---it cannot appropriately distinguish overlapping objects. Such a mechanism poses problems for accurate maize tassels detection because overlaps between different tassels are common patterns in the field. We have to ask: is the object detection the best way to count maize tassels? From a point of view of Computer Vision, the objective of object detection is to localise individual instances and output their corresponding bounding boxes. Since the locations of objects are identified, it is easy to derive the number of instances. However, the number of instances actually has nothing to do with the location. If one only cares about estimating the total number of instances, the problem is another important research topic in Computer Vision---object counting. In this paper, we show that it is better to formulate the task of maize tassels counting as a typical counting problem, rather than a detection one. In fact, object detection is generally more difficult to solve than object counting.

\begin{figure*}[h!]
	\centering
	\includegraphics[width=0.9\textwidth]{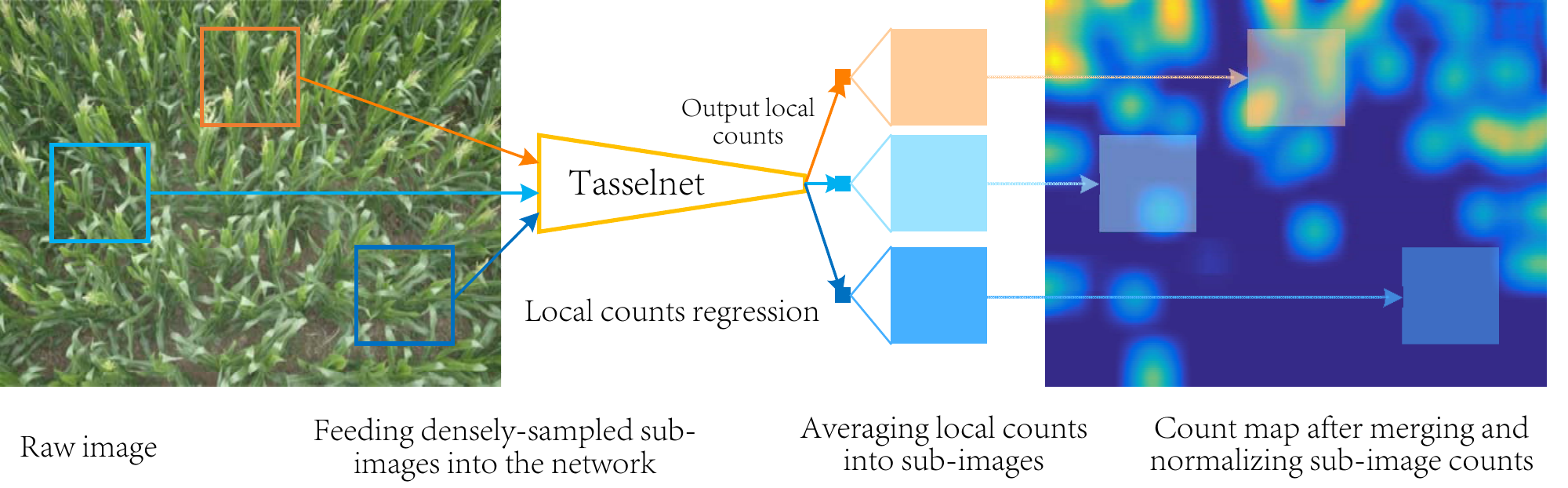}
	\caption{The main technical pipeline of in-field counting of maize tassels. Sub-images are first densely sampled from a raw field image. Each sub-image will be fed into our Tasselnet to regress a local count associating with the sub-image. After merging and normalizing all local counts, a count map for the field image can be acquired. The raw image count can thus be computed by integrating the count map.}
	\label{fig:pipeline}
\end{figure*}

Nevertheless, object counting remains a known challenging task~\cite{minervini2015image,ali2013modeling}, in both Plant Science and Computer Vision communities. Three sessions of \emph{Leaf Counting Challenge} have been held in conjunction with the \emph{Computer Vision Problems in Plant Phenotyping} workshops (CVPPP2014\footnote{\url{https://www.plant-phenotyping.org/CVPPP2014}}/CVPPP2015\footnote{\url{https://www.plant-phenotyping.org/CVPPP2015}}/ CVPPP2017\footnote{\url{https://www.plant-phenotyping.org/CVPPP2017}}), expecting to showcase visual challenges for plant phenotyping. Many efforts are also made in recent years in Computer Vision to improve the counting precision of crowds~\cite{chan2008privacy,zhang2015cross}, cells~\cite{vlaz2010denlearn,xie2016microscopy}, cars~\cite{arteta2014interactive,onoro2016towards}, and animals~\cite{arteta2016counting}. However, little attention has been paid to plants-related counting tasks. To our knowledge, only two published papers considered counting problems relating to plants.~\cite{giuffrida2015learning} proposed a learning-based approach to count leaves in rosette plants.~\cite{rahnemoonfar2017deep} presented a deep simulated learning approach to count tomato images. A limitation is that both papers only report their results on potted plants, which is far different from field-based scenarios. In contrast, our experiments use images captured exactly under unconstrained in-field environment, leading to a more challenging situation and a more reasonable experimental evaluation.

According to the taxonomy of~\cite{loy2013crowd}, existing object counting approaches can be classified into three categories: counting by clustering, counting by detection, and counting by regression. The counting-by-clustering approaches often rely on the extraction of motion features (see~\cite{rabaud2006counting} for example), which is not applicable to the plants because the motion of plants is almost unobservable within limited time. In addition, the counting-by-detection approaches~\cite{li2008estimating,dollar2012pedestrian} tend to suffer in crowded scenes with significant occlusions, so this type of method is also not a good choice for our problem. In fact, the transductive principle suggests never to solve a harder problem than the target application necessitates~\cite{vapnik1998statistical}. As a consequence, recent counting-by-regression models~\cite{chan2008privacy,vlaz2010denlearn,arteta2014interactive} have demonstrated that it is indeed unnecessary to detect or segment individual instances when estimating their counts. In particular, the key component of modern counting-by-regression approaches is the introduction of the density map by Lempitsky and Zisserman~\cite{vlaz2010denlearn}. Objects in an image are described by a density map given dot annotations. During the prediction, each object will be assigned a density taking up 1, so the total number of objects can be reflected by summing over the whole density map. Overlapping objects are naturally taken into account in this paradigm.

To better address aforementioned challenges, we follow the idea of counting by regression and propose in this paper a deep convolutional neural network~\cite{Krizhevsky2012} (CNN)-based approach for maize tassels counting, which is referred to \textit{Tasselnet}. Deep networks are famous due to their excellent non-linear modelling ability and large model capacity, which is important for capturing diverse and complex visual patterns in the field. Notice that plants are like self-changing systems, the physical size of maize tassels in images vary significantly over time. This is what makes the problem of maize tassels counting different from other conventional counting problems in Computer Vision (the physical size of pedestrians, cells or cars in images remains unchanged or almost identical), and consequently, renders difficulties to describe the density map of maize tassels. To address this, in contrast to~\cite{vlaz2010denlearn} and~\cite{onoro2016towards} that either regress the global density map or the local density map, we propose to regress the local count computed from the density map. After merging and normalizing all local counts, our model outputs a count map similar to the ground-truth density map. The final count of maize tassels is computed by summing over the whole count map. Fig.~\ref{fig:pipeline} illustrates our main technical pipeline.

\begin{figure*}[!h]
	\centering
	\includegraphics[width=0.7\textwidth]{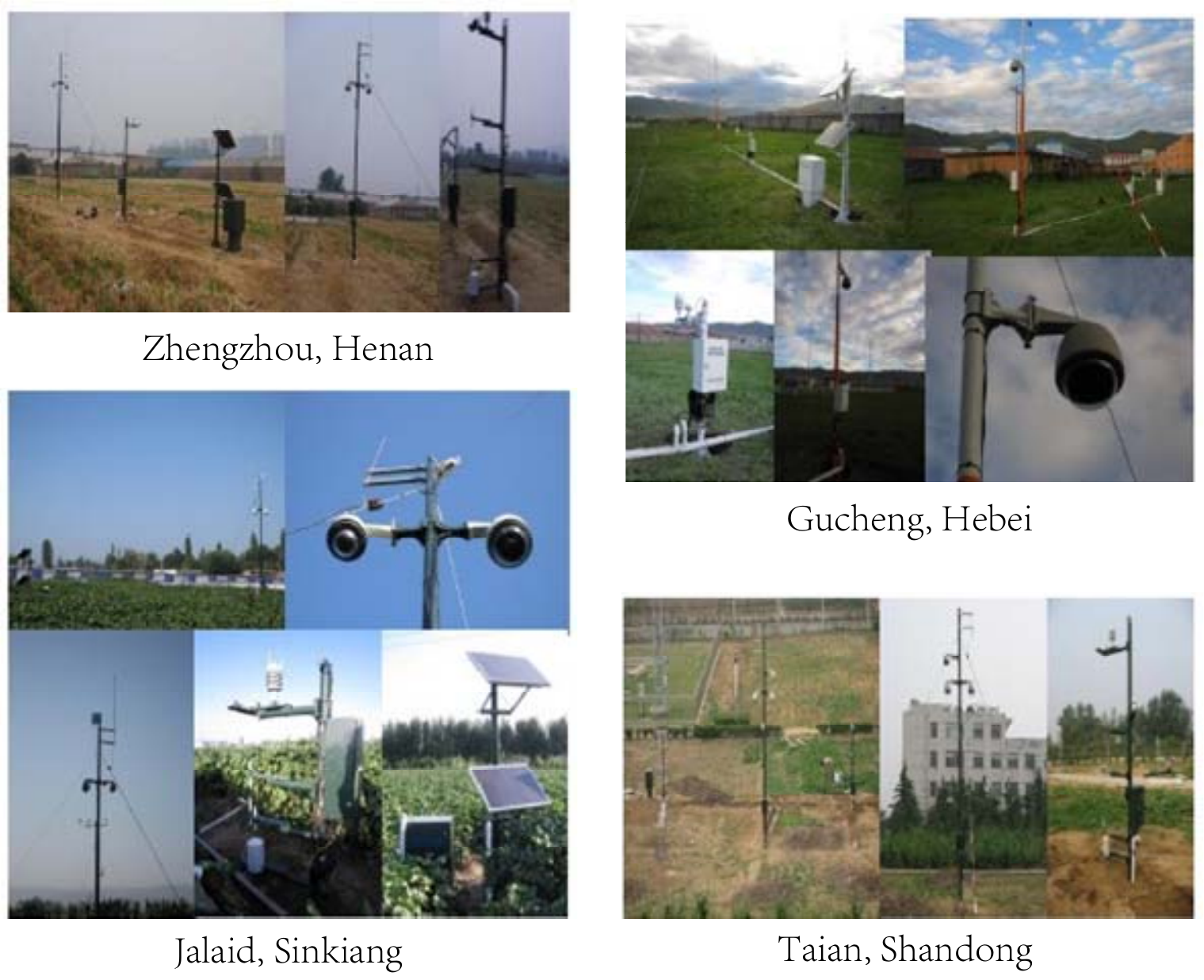}
	\caption{Image acquisition devices in the maize field. Our devices are currently set up in four different places.}
	\label{fig:device}
\end{figure*}

To validate the effectiveness of the proposed approach, a novel \emph{Maize Tassel Counting (MTC)} dataset is constructed and will be released together with this paper. Our MTC dataset contains 361 images chosen from 16 image sequences. These sequences are collected from 2010 to 2015, covering 4 different experimental fields across China. All challenges described in Fig.~\ref{fig:challenges} are involved in this dataset. The number of maize tassels in images varies between 0 and around 100. Following the standard annotation used in objection counting problems~\cite{vlaz2010denlearn}, a single dot is manually assigned for each maize tassel. We hope such a dataset could be used as a benchmark for evaluating in-field counting approaches and could draw attention from practitioners working in this area to attach importance to these in-field challenges.

Extensive evaluations are performed on the MTC dataset. Experimental results demonstrate that Tasselnet outperforms other state-of-the-art methods and significantly reduces the counting errors by large margins. Moreover, based on the experimental results, we also suggest several good practices for in-field counting problems.

The contributions of this paper are multi-fold:
\begin{itemize}
	\item A novel counting problem of maize tassels whose sizes are self-changing over time. To the best of our knowledge, this is the first time that a plant-related counting problem is considered under unconstrained field conditions;
	\item A challenging MTC dataset with 361 field images and corresponding manually-labelled dotted annotations;
	\item Tasselnet: an effective deep CNN-based solution for in-field counting of maize tassels via local counts regression.
\end{itemize}

\section*{Methods}

\subsection*{Experimental fields and imaging devices}
16 independent time-series image sequences are collected from four different experimental fields across China between 2010 and 2015. Four experimental fields are located in Zhengzhou, Henan Province, China, Taian, Shandong Province, China, Gucheng, Hebei Province, China, and Jalaid, Sinkiang Autonomous Region, China, respectively. Six cultivars of maize plants are involved, including Jundan No.20, Nongda No.108, Wuyue No.3, Zhengdan No.32, Jidan No.20, and Tianlong No.9. Fig.~\ref{fig:device} shows the experimental fields and imaging devices. The main components of the imaging device include a high-resolution CCD digital camera (E450 Olympus), a low-resolution monitoring device, a 3G wireless data transmission system, as well as several solar panels used for power supply. When an image is captured, it will be transmitted into a remote server, and then users can access the image data. Readers can refer to~\cite{lu2016toward} for a detailed introduction of our imaging device. The focal length of the camera is fixed to 16mm. Images were taken every one hour from 9:00 to 16:00 from a
five-meters-height vertical view (four meters for Gucheng sequences). The original image resolutions are 3648$\times$2736 pixels for Zhengzhou and Taian sequences, 4272$\times$2848 pixels for Gucheng sequences, and 3456$\times$2304 pixels for Jalaid sequences.

\subsection*{Maize tassels counting dataset}
Given 16 independent time series image sequences, images captured from the tasselling stage to the flowering stage are considered in our MTC dataset. In particular, according to the variability each sequence presents, 8$\sim$45 images are manually chosen from each sequence. If extrinsic conditions, such as weather conditions or the wind, change dramatically, more images will be chosen in one day, otherwise only 1 or 2 images are chosen. Such a sampling strategy is used with the motivation to avoid repetitive samples as much as possible, because images captured in one day usually do not exhibit many variations. However, the ability to model various data variabilities is much more important than blindly fitting a large number of repetitive samples for an effective computer vision approach. Thus, 361 field images in all are chosen to construct the MTC dataset. The MTC dataset is divided into the training set, validation set and test set. The training set and validation set share the same image sequences, while the test set uses different image sequences to enable a reasonable evaluation. Such an intentional setting is motivated by the fact that images in one sequence are often highly correlated, it is thus inappropriate to place them into both the training and test stages. Table~\ref{tab:dataset} summarises the information of the MTC dataset. Overall, we have 186 images for training and validation and 175 images for test.

\begin{figure*}[h!]
	\centering
	\includegraphics[width=0.95\textwidth]{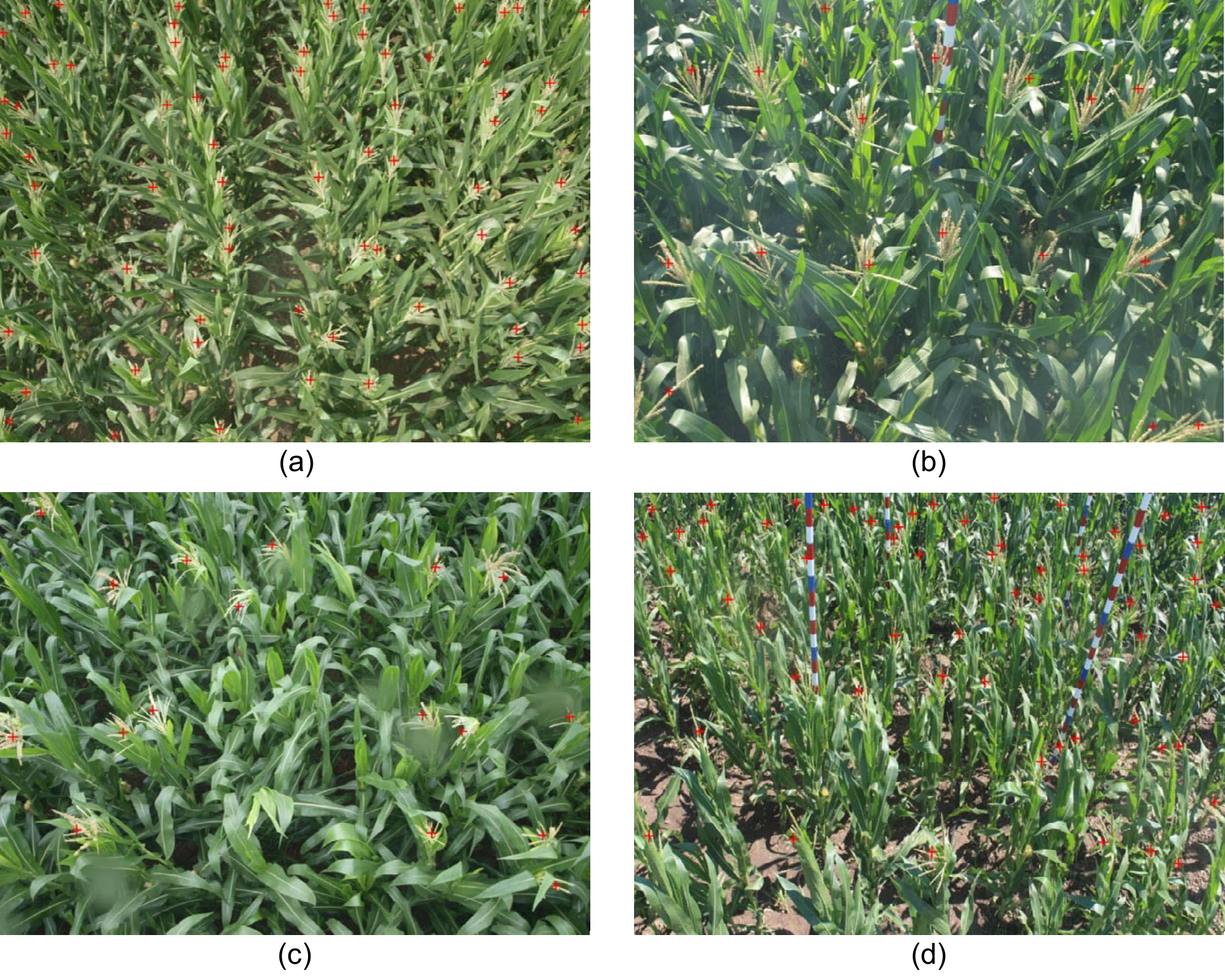}
	\caption{Example images in the MTC dataset with dotted annotations. Images are from the (a) Zhengzhou2010, (b) Gucheng2012, (c) Taian2011\_1 and (d) Jalaid2015\_1 sequences, respectively.}
	\label{fig:dataset}
\end{figure*}

We also follow the standard annotation paradigm that manually provides each tassel with a dot annotation~\cite{vlaz2010denlearn}. Indeed, dotting is regarded as a natural way to count for humans. It not only gives the raw counts of the image but also proffer the information how objects spatially distribute. Fig.~\ref{fig:dataset} shows four example images with the dotted annotations.

\begin{table}[!t]
	\centering
	\caption{Training set (train), validation set (val) and test set (test) settings of the MTC dataset. $Num$ refers to the number of images in each sequence.}
	\label{tab:dataset}
	\renewcommand\arraystretch{1.25}
	\addtolength{\tabcolsep}{-1pt}
	\begin{tabular}{l|clccc}
		\hline
		Sequence & $Num$ & Cultivar & train & val & test\\
		\hline
		Zhengzhou2010 & 37 & Jundan No.20    & \checkmark & \checkmark & \\
		Zhengzhou2011 & 24 & Jundan No.20    &            &            & \checkmark\\
		Zhengzhou2012 & 22 & Zhengdan No.958 & \checkmark & \checkmark & \\
		Taian2010\_1  & 30 & Wuyue No.3      & \checkmark & \checkmark & \\
		Taian2010\_2  & 32 & Wuyue No.3      &            &            & \checkmark\\
		Taian2011\_1  & 21 & Nongda No.108   & \checkmark & \checkmark & \\
		Taian2011\_2  & 19 & Nongda No.108   &            &            & \checkmark\\
		Taian2012\_1  & 41 & Zhengdan No.958 & \checkmark & \checkmark & \\
		Taian2012\_2  & 23 & Zhengdan No.958 &            &            & \checkmark\\
		Taian2013\_1  & 8  & Zhengdan No.958 & \checkmark & \checkmark & \\
		Taian2013\_2  & 8  & Zhengdan No.958 &            &            & \checkmark\\
		Gucheng2012   & 15 & Jidan No.32     & \checkmark & \checkmark & \\
		Gucheng2014   & 45 & Zhengdan No.958 &            &            & \checkmark\\
		Jalaid2015\_1 & 12 & Tianlong No.9   & \checkmark & \checkmark & \\
		Jalaid2015\_2 & 12 & Tianlong No.9   &            &            & \checkmark\\
		Jalaid2015\_3 & 12 & Tianlong No.9   &            &            & \checkmark\\
		\hline
	\end{tabular}
\end{table}

\subsection*{Local counts regression network}
In this section we describe our proposed local counts regression network and show how to use it to address effectively the in-field counting problem of maize tassels.

The high-level idea of counting by regression is simple: given an image $I$ and a regression target $T$, the goal is to seek some kind of regression function $F$ so that $T\approx F(I)$. Standard solutions are to regress explicitly the raw counts in an image~\cite{chan2008privacy} ($T$ is the global counts) or to regress implicitly the density map of an image~\cite{vlaz2010denlearn} ($T$ becomes a density map, and the counts can be acquired by integrating over the entire density map). However, as what we will show in our experiments, both solutions are not effective for maize tassels counting. The reason may boil down to the heterogeneity of maize tassels. As shown in Fig.~\ref{fig:dataset}, maize tassels exhibit uncertain poses and varying sizes, making them hard to be described by only a global image representation or a density map given only dotted annotations. Indeed, this is what makes maize tassels counting different from other standard counting problems.

\begin{figure}[h!]
	\centering
	\includegraphics[width=0.95\linewidth]{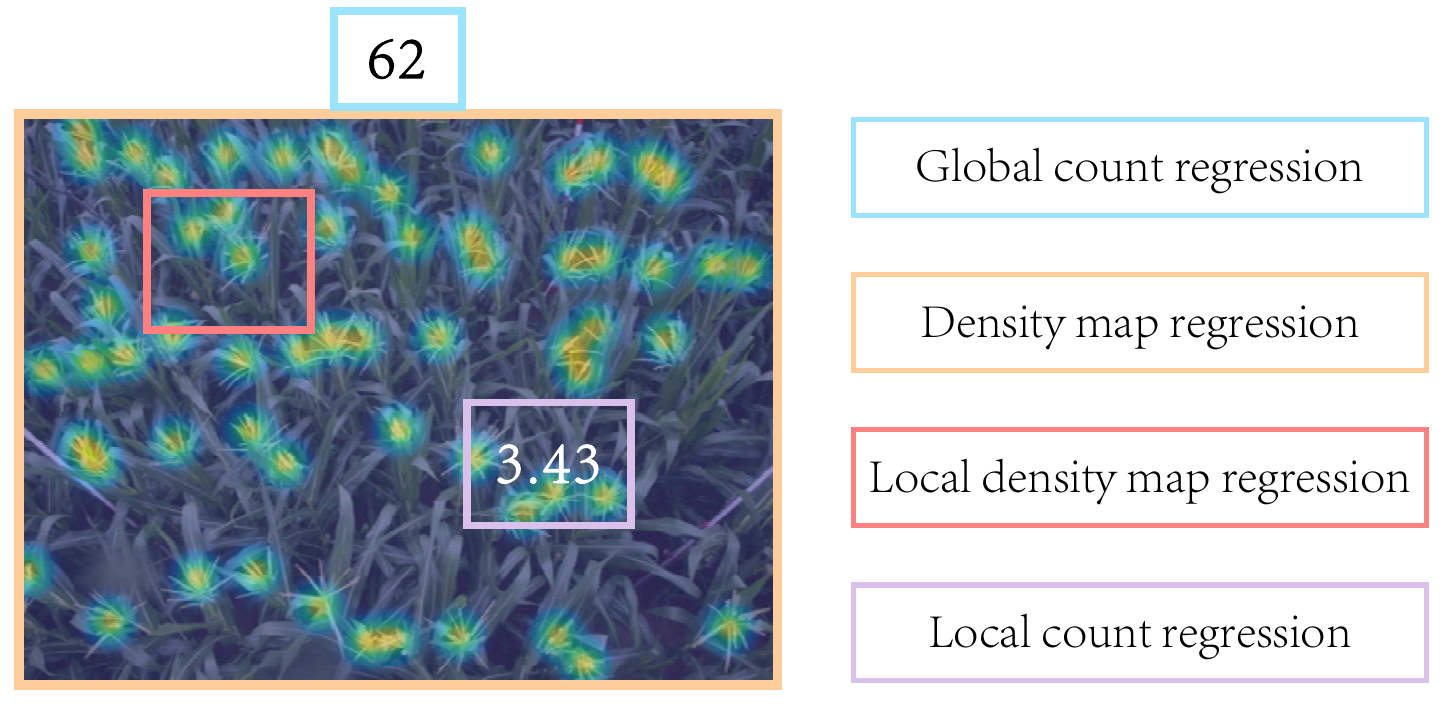}
	\caption[width=\textwidth]{Conceptual differences of different regression targets. The global count regression directly regresses the number of image counts in an image. (Local) density map regression treats the two-dimensional (local) density map as the regression target. Our proposed local count regression regresses the local count computed from the local density map. (Best viewed in colour.)}
	\label{fig:reg_diff}
\end{figure}

Inspired by a recent idea of redundant counting~\cite{cohen2017count}, we propose to regress the local counts $T_l$ to address the counting problem of maize tassels. $T_l$ refers to the object counts within a small sub-image $I_s$. The proposed local regression has several benefits: i) Local characteristics are easier to be modelled than the global ones; ii) By regressing the local counts, we avoid the hard problem of dense per-pixel learning (compared to estimating the local density map); iii) By sampling small image patches, we can have access to a large number of training data, allowing us to train a high-capacity model. In particular, we consider the regression function $F$ should be powerful enough so that it can appropriately capture those heterogeneous in-field variations. Inspired by the recent success of deep convolutional neural networks (CNNs) in visual recognition~\cite{Krizhevsky2012,Simonyan14verydeep}, we choose to formulate $F$ in a deep CNN-based framework. The goal is thus to recover $T_l$ with a set of non-linear transformations $F$ given $I_s$, i.e., $T_l\approx F(I_s)$. Fig.~\ref{fig:reg_diff} compares the conceptual difference of different regression goals. During the prediction, sub-images are densely sampled from a test image, and $F$ will assign a local count to each sub-image patch. The final count of the image can be recovered by aggregating all sub-image counts into a count map with the same size of the test image, and a per-pixel normalisation step is performed to each pixel by dividing the number of sub-images that cast a prediction in it.

\paragraph{Regression target}

\begin{figure}[!t]
	\includegraphics[width=0.95\linewidth]{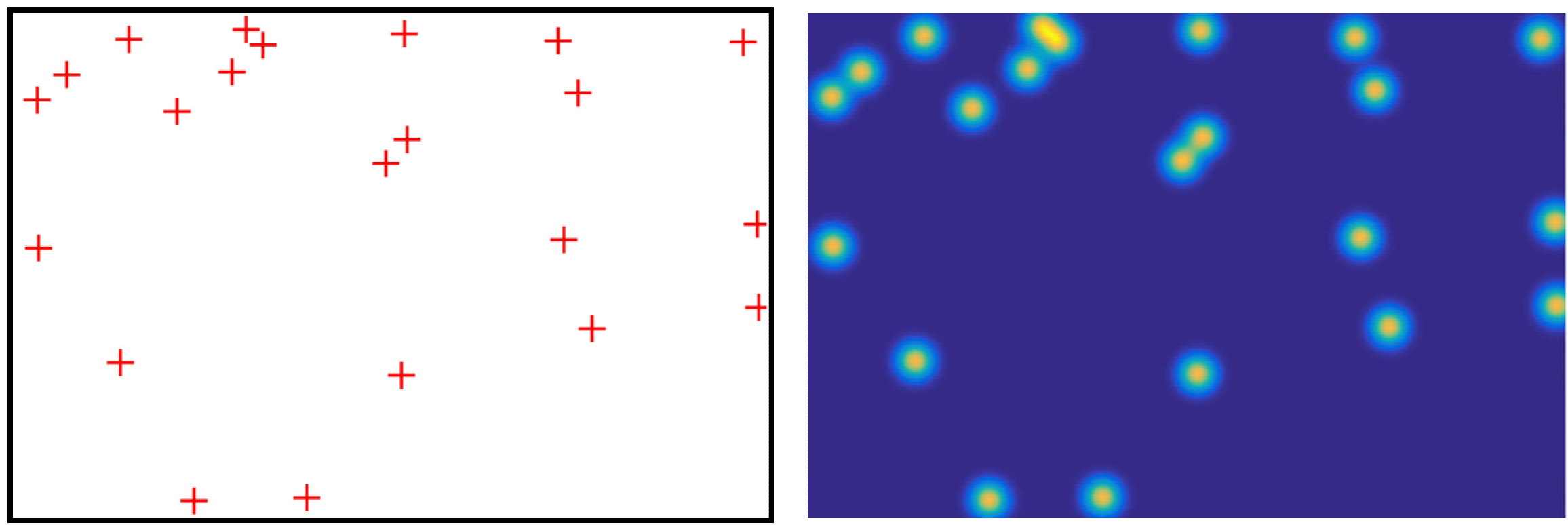}
	\caption{An example of manually-annotated dot image (left) and its corresponding ground truth density map (right).}
	\label{fig:densitymap}
\end{figure}

Different regression targets imply different regression strategies, so how to define the regression target is the first and the most important step. In this paper, we first follow the standard way that generates the ground truth density by placing a Gaussian at each dot annotation~\cite{vlaz2010denlearn}. Formally, given a ground-truth dot image $Y$, a density map $D$ can be defined as $D=G*Y$, where $G$ denotes a two-dimensional Gaussian kernel parametrised by $\sigma$, and $*$ indicates the convolution operation. Obviously, $D$ is generated by performing Gaussian smoothing on $Y$. Fig.~\ref{fig:densitymap} shows an example of $D$ given corresponding dotted annotations. It is worth noting that the summation of the density map is a decimal. The reason is that, when dots are close to the image boundary, their Gaussian probability will be partly outside the image, but this definition naturally takes a fraction of an objection into account.

However, in contrast to~\cite{vlaz2010denlearn,onoro2016towards}, we do not regard $D$ as our regression target (we will show later in our experiments that the density map is too harsh as the regression target) but use the local counts integrated from the density map. If let $D(x,y)$ be the pixel-level count of $D$ at the location $(x,y)$, then the regression target of local counts $T_l^i$ for the $i$-th sub-image $I_s^i$ can be defined:
\begin{equation}
T_l^i=\sum_{(x,y)\in S_i}D(x,y)\,,
\end{equation}
where $S_i$ denotes the set of pixel locations of $I_s^i$.

\begin{figure*}[h!]
	\centering
	\includegraphics[width=0.95\textwidth]{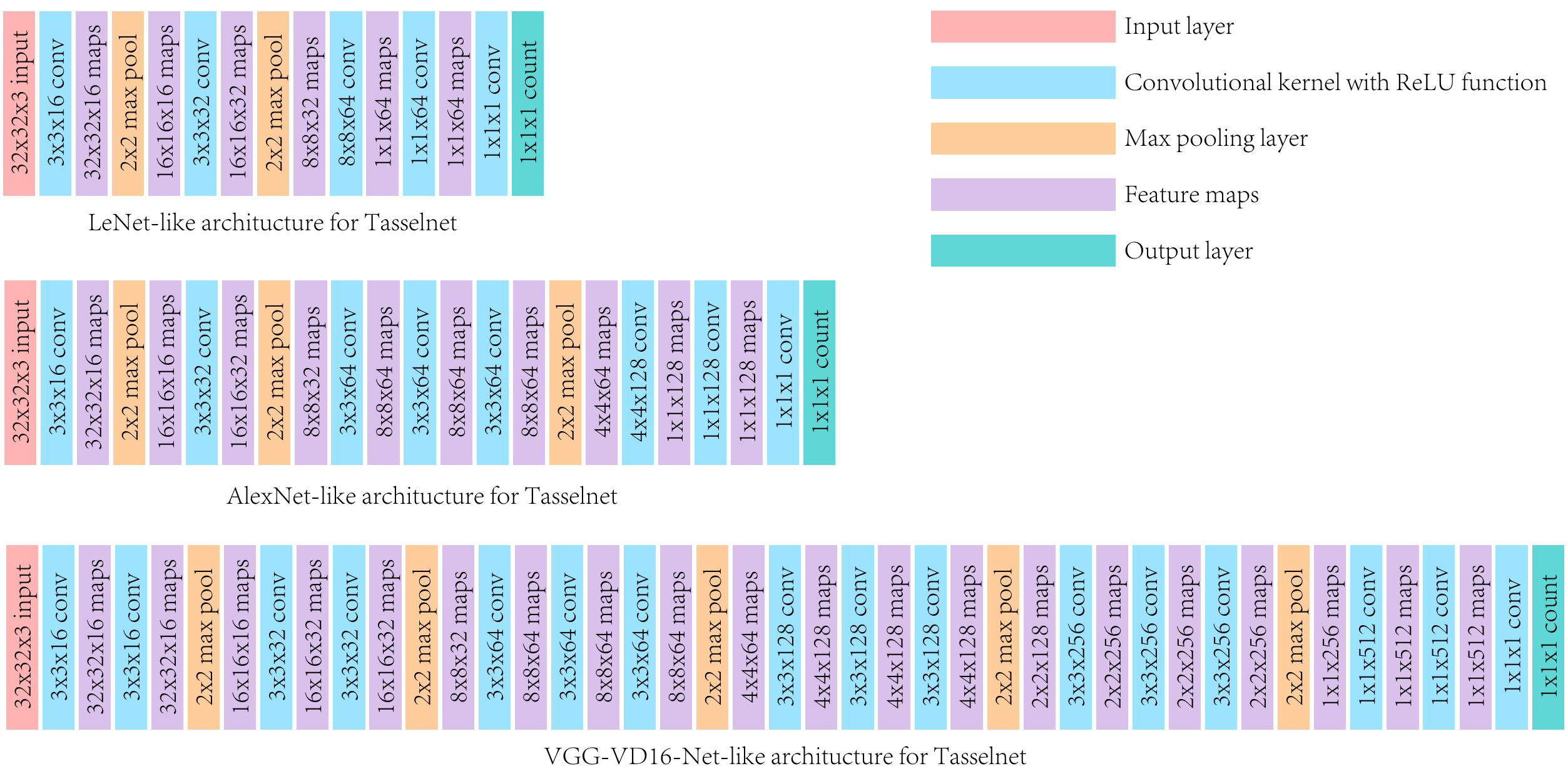}
	\caption{Three typical CNN architectures used in Tasselnet.}
	\label{fig:network}
\end{figure*}

\paragraph{Network architecture}

The network architecture closely relates to to the model capacity, and the model capacity is also a key factor that affects the counting performance. Motivated by the leading role of CNNs in Computer Vision, in this paper we evaluate three typical CNNs architectures: a low-capacity 4-layer model identical to the seminal LeNet architecture~\cite{lecun1998gradient}, a medium-capacity 7-layer model similar to the AlexNet architecture~\cite{Krizhevsky2012}, as well as a high-capacity 16-layer model sharing the same spirit of the VGG-VG16-Net~\cite{Simonyan14verydeep}.

We follow the modern CNN design principle used in~\cite{Simonyan14verydeep}: adopting only small $3\times 3$ convolution kernels with 1-pixel padding to preserve the size of the tensor, doubling the number of feature maps in the higher layers to compensate the loss of the spatial information after the max pooling operation, synthesizing learnt features with two extra fully-connected layers, and using the ReLU function after each convolutional/full-connected layer. Fig.~\ref{fig:network} shows three architectures with a basic input size of $32\times 32$ sub-image. The number of parameters within the LeNet-like, AlexNet-like and VGG-VD16-Net-like architectures are about $1.4\times 10^5$, $2.5\times 10^5$ and $2.4\times 10^6$, respectively.

\paragraph{Loss function}

The learning of the regression network should be driven by a loss function. In this paper, we evaluate three typical loss functions used in regression problems. They are $\ell_1$ loss, $\ell_2$ loss, and \textit{Huber} loss. $\ell_1$ loss and $\ell_2$ loss take the form
\begin{equation}
L_1(\theta)=\frac{1}{M}\sum_{i=1}^M\|F(I_c^i|\theta)-T_l^i\|_1\,,
\end{equation}
\begin{equation}
L_2(\theta)=\frac{1}{M}\sum_{i=1}^M\|F(I_c^i|\theta)-T_l^i\|_2^2\,,
\end{equation}
where $L1$ and $L2$ denote the $\ell_1$ and $\ell_2$ loss functions, respectively. $\theta$ is the network parameter, and $M$ is the number of training sub-images. Empirically, $\ell_1$ loss is considered more robust to noise than $\ell_2$ loss. Apart from these two standard choices, another widely-used loss function in robust regression is the \textit{Huber} loss, which is defined as
\begin{equation}
L_\delta(a^i)=\begin{cases}
	\frac{1}{2M}\sum_{i=1}^M\|a^i\|^2_2\,, &\text{if }|a^i|\leq\delta\\
	\frac{1}{M}\sum_{i=1}^M\delta|a^i|-\frac{1}{2}\delta^2\,, &\text{otherwise}
	\end{cases}\,,
\end{equation}
$a^i=F(I_c^i|\theta)-T_l^i$, and $\delta$ is a user-defined constant. \textit{Huber} loss can be viewed as an integration of $\ell_1$ and $\ell_2$ losses. We will show later in our experiments that $\ell_1$ loss is the most effective one for maize tassels counting.

\paragraph{Merging and normalizing sub-image counts}
During the prediction, Tasselnet will scan the image in a sliding window manner with a stride of $s_e$. Each window corresponds to a sub-image with size of $r\times r$. For each sub-image, Tasselnet regresses a local count indicating the number of tassels within the sub-image. Assume that $K$ sub-images in all are processed. Since each sub-image may be counted multiple times due to the densely-sampled mechanism, the final count of maize tassels cannot be directly computed by simply summing over all $K$ local counts. To address this, here we develop a merging strategy to map $K$ local counts back to the original test image. Assume that the $k$-th sub-image count is $c_k$, we average $c_k$ into every pixel of the $k$-th sub-image, so the count of each pixel takes up $\frac{c_k}{r^2}$ (the sum of pixel-level counts still equals to $c_k$). In this way, a count map $C$ with the same resolution of the test image can be consequently constructed by mapping the $r\times r$ local count maps back to the same location where the sub-image is sampled. Fig.~\ref{fig:pipeline} illustrates this process. Finally, by constructing a normalisation image $P$ that records how many times each pixel is counted, the final count of the image $c$ can be computed as
\begin{equation}\label{eq:count}
c=\sum_{x,y}\frac{C(x,y)}{P(x,y)}\,,
\end{equation}
where $C(x,y)$ and $P(x,y)$ denote the value of $C$ and $P$ at the location $(x,y)$.

\paragraph{Implementation and learning details}
We implement Tasselnet based on \texttt{MatConvNet}~\cite{Vedaldi2015conv}. Original high-resolution images are resized to their $1/8$ sizes to reduce computational burden. During training, we densely crop $r\times r$ sub-images with a stride of $s_r$ from 186 images belonging to the training and validation sequences of MTC dataset. We perform a random shuffling of these sub-images, 90\% sub-images are used for training, and the rest for validation. Before feeding the image samples into the network, each sub-image is preprocessed by mean subtraction (the mean is computed from the training subset).

The parameters of the convolution kernels are initialised with the \textit{improved Xaiver} method~\cite{he2015delving}. The standard stochastic gradient descent is used to optimise the network parameters. The learning rate is initially set to 0.01, and is decreased by a factor of 10 after 5 epochs and further decreased by a factor of 10 after another 10 epochs. Thus, we train Tasselnet for 25 epochs in all. To allow the gradient to back-propagate easily from the output layer to the input layer, we add a batch normalisation layer~\cite{ioffe2015batch} after each convolutional layer before ReLU. The training time of Tasselnet varies from half a day to 2 days depending on the number of training samples and the network architecture used. The prediction time for each image takes about 2.5 seconds (Matlab 2016a, OS: Ubuntu 14.04 64-bit, CPU: Intel E5-2630 2.40GHz, GPU: Nvidia GeForce GTX TITAN X, RAM: 64 GB).

Table~\ref{tab:parameters} summarises the default parameters used in our experiments. When $s_r=8$, 355,473 and 31,167 sub-images are densely sampled and ready for training and validation, respectively.

\begin{table}[!t]
	\centering
	\caption{Default parameters setting used in our experiments.}
	\label{tab:parameters}
	\renewcommand\arraystretch{1.25}
	\addtolength{\tabcolsep}{-1pt}
	\begin{tabular}{l|ll}
		\hline
		Parameter & Remark & Value\\
		\hline
		& network architecture & AlexNet-like\\
		& loss function & $\ell_1$\\
		\hline
		$r$ & sub-image size & 32 \\
		$s_r$ & sampling stride during training & $r/4$\\
		$s_e$ & sampling stride during prediction & $r/4$\\
		$\sigma$ & Gaussian kernel parameter & 8\\
		\hline
	\end{tabular}
\end{table}

\begin{figure*}[!t]
	\centering
	\includegraphics[width=0.96\textwidth]{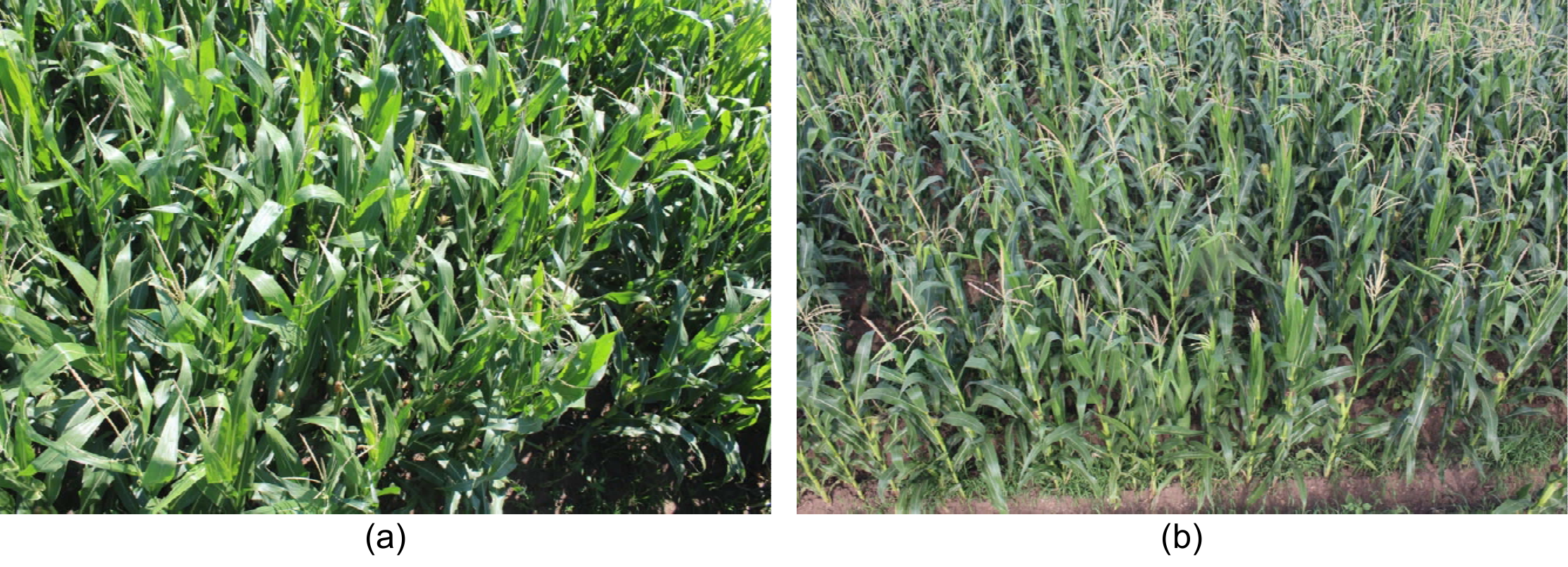}
	\caption{Two example images from the Jalaid2015\_2 and Jalaid2015\_3 sequences. Images in two sequences exhibit dramatic illumination variations, dazzling visual characteristics, as well as extremely crowded distributions, which renders great difficulties for counting even for a human expert.}
	\label{fig:challenging_seq}
\end{figure*}

\section*{Results and discussion}
We evaluate the effectiveness of Tasselnet on each test sequence of MTC dataset, respectively. It is worth noting that Jalaid2015\_2 and Jalaid2015\_3 are two very challenging sequences. As shown in Fig.~\ref{fig:challenging_seq}, images in the Jalaid2015\_2 sequence suffer from dramatic illumination variations (Jalaid locates in a high-latitude area), and maize tassels in the Jalaid2015\_2 sequence exhibit extremely crowded distributions. Extensive experiments are conducted to investigate key factors that affect the counting performance and to compare Tasselnet against other state-of-the-art approaches. Based on the experimental results, we also suggest several good practices for practitioners working on in-filed counting problems.

\subsection*{Evaluation metric}
The mean absolute error (MAE) and the mean squared error (MSE) are used as the evaluation metrics to assess the counting performance. They take the form
\begin{equation}
MAE=\frac{1}{N}\sum_1^N|t_i-c_i|\,,
\end{equation}
\begin{equation}
MSE=\sqrt{\frac{1}{N}\sum_1^N(t_i-c_i)^2}\,,
\end{equation}
where $N$ denotes the number of test images, $t_i$ is the ground truth count for the $i$-th image (computed by summing over the whole density map), and $c_i$ is the inferred image count for the $i$-th image (computed as per Eq.~\eqref{eq:count}). $MAE$ quantifies the accuracy of the estimates, and $MSE$ assesses the robustness of the estimates. The lower these two measures are, the better the counting performance is.

\begin{table*}[!t] \footnotesize
	\centering
	\caption{Comparison of different network architectures for maize tassels counting on the MTC dataset. The lowest error is boldfaced.}
	\label{tab:architecture}
	\renewcommand\arraystretch{1.25}
	\addtolength{\tabcolsep}{-3.5pt}
	\begin{tabular}{l|cc|cc|cc|cc|cc|cc|cc|cc|cc}
		\hline
		Network                & \multicolumn{16}{c|}{Sequences} & \multicolumn{2}{c}{Overall}\\
		\hline
		& \multicolumn{2}{c|}{\scriptsize Zhengzhou2011}  & \multicolumn{2}{c|}{\scriptsize Taian2010\_2} & \multicolumn{2}{c|}{\scriptsize Taian2011\_2} & \multicolumn{2}{c|}{ \scriptsize Taian2012\_2} & \multicolumn{2}{c|}{\scriptsize Taian2013\_2} & \multicolumn{2}{c|}{\scriptsize Gucheng2014} & \multicolumn{2}{c|}{\scriptsize Jalaid2015\_2} & \multicolumn{2}{c|}{\scriptsize Jalaid2015\_3} & \\
		& MAE  & MSE  & MAE  & MSE  & MAE  & MSE  & MAE  & MSE  & MAE  & MSE  & MAE  & MSE  & MAE  & MSE  & MAE  & MSE  & MAE  & MSE \\
		LeNet     & 4.4 & 5.4 & 6.3 & 8.0 & 2.9 & 3.7 & 6.4 & 7.9 & 4.9 & 5.8 & \textbf{3.8} & \textbf{5.0} & 16.3 & 17.0 & 28.7 & 33.0 & 7.2 & 11.3\\
		AlexNet   & 4.9 & 6.1 & \textbf{5.2} & \textbf{6.6} & \textbf{2.5} & \textbf{2.9} & \textbf{4.8} & \textbf{5.8} & \textbf{4.0} & \textbf{5.0} & 5.3 & 6.5 & 16.0 & 16.6 & \textbf{20.7} & 25.2 & \textbf{6.6} & \textbf{9.6}\\
		VGG-VD16-Net  & \textbf{2.1} & \textbf{2.7} & 10.6 & 12.4 & 13.1 & 15.9 & 5.5 & 10.0 & 4.3 & 5.4 & 10.0 & 11.3 & 10.7 & 11.2 & 20.8 & \textbf{24.9} & 9.3 & 12.4\\
		\hline
	\end{tabular}
\end{table*}

\begin{table*}[!t] \footnotesize
	\centering
	\caption{Counting performance with different number of training samples ($N_{train}$) on the MTC dataset. The lowest error is boldfaced.}
	\label{tab:num_subimages}
	\renewcommand\arraystretch{1.25}
	\addtolength{\tabcolsep}{-3.5pt}
	\begin{tabular}{l|cc|cc|cc|cc|cc|cc|cc|cc|cc}
		\hline
		$N_{train}$ & \multicolumn{16}{c|}{Sequences} & \multicolumn{2}{c}{Overall}\\
		\hline
		& \multicolumn{2}{c|}{\scriptsize Zhengzhou2011}  & \multicolumn{2}{c|}{\scriptsize Taian2010\_2} & \multicolumn{2}{c|}{\scriptsize Taian2011\_2} & \multicolumn{2}{c|}{ \scriptsize Taian2012\_2} & \multicolumn{2}{c|}{\scriptsize Taian2013\_2} & \multicolumn{2}{c|}{\scriptsize Gucheng2014} & \multicolumn{2}{c|}{\scriptsize Jalaid2015\_2} & \multicolumn{2}{c|}{\scriptsize Jalaid2015\_3} & \\
		& MAE  & MSE  & MAE  & MSE  & MAE  & MSE  & MAE  & MSE  & MAE  & MSE  & MAE  & MSE  & MAE  & MSE  & MAE  & MSE  & MAE  & MSE \\
		$2.37\times10^4$   & 6.3 & 7.9 & 8.5 & 11.0 & 5.1 & 6.1 & 7.5 & 9.8 & 8.0 & 11.6 & 6.9 & 8.5 & 16.5 & 18.4 & 33.0 & 39.4 & 9.5 & 14.2\\
		$9.13\times10^4$   & 4.9 & 5.9 & 6.9 & 8.6 & 4.2 & 5.3 & 6.5 & 7.9 & 5.7 & 7.0 & 5.5 & 7.2 & 18.4 & 19.5 & 34.2 & 40.0 & 8.5 & 13.4\\
		$3.56\times10^5$   & 4.9 & 6.1 & 5.2 & 6.6 & \textbf{2.5} & \textbf{2.9} & 4.8 & 5.8 & \textbf{4.0} & \textbf{5.0} & 5.3 & 6.5 & \textbf{16.0} & \textbf{16.6} & \textbf{20.7} & \textbf{25.2} & 6.6 & \textbf{9.6}\\
		$1.41\times10^6$   & \textbf{3.9} & \textbf{4.8} & \textbf{4.6} & \textbf{5.8} & 2.7 & 3.0 & \textbf{3.8} & \textbf{4.4} & 4.2 & 5.2 & \textbf{4.3} & \textbf{5.3} & \textbf{16.0} & 16.7 & 29.1 & 33.4 & \textbf{6.5} & 10.8\\
		\hline
	\end{tabular}
\end{table*}

\subsection*{Choices of different network architectures, number of training samples, loss functions, Gaussian kernel parameters, and sub-image sizes}
Here we perform extensive evaluations to justify our design choices. Notice that, in a principal way, the inclusion of specific design choices should be justified on the validation set. However, since we enforce the test set to be different sequences, the validation set thus exhibits a substantially different data distribution. Validating our design choices based on the validation set seems suboptimal to the test set. Instead, as a preliminary study, we direct report the counting performance on the test set to see how the variations of these design choices affect the final counting performance. Although with a little abuse, we demonstrate later that the performance of Tasselnet with any design choice shows a notable improvement over other baseline approaches by large margins. Below we follow the default parameters setting unless a specific design choice is declared.

\begin{figure}[!t]
	\centering
	\includegraphics[width=0.75\linewidth]{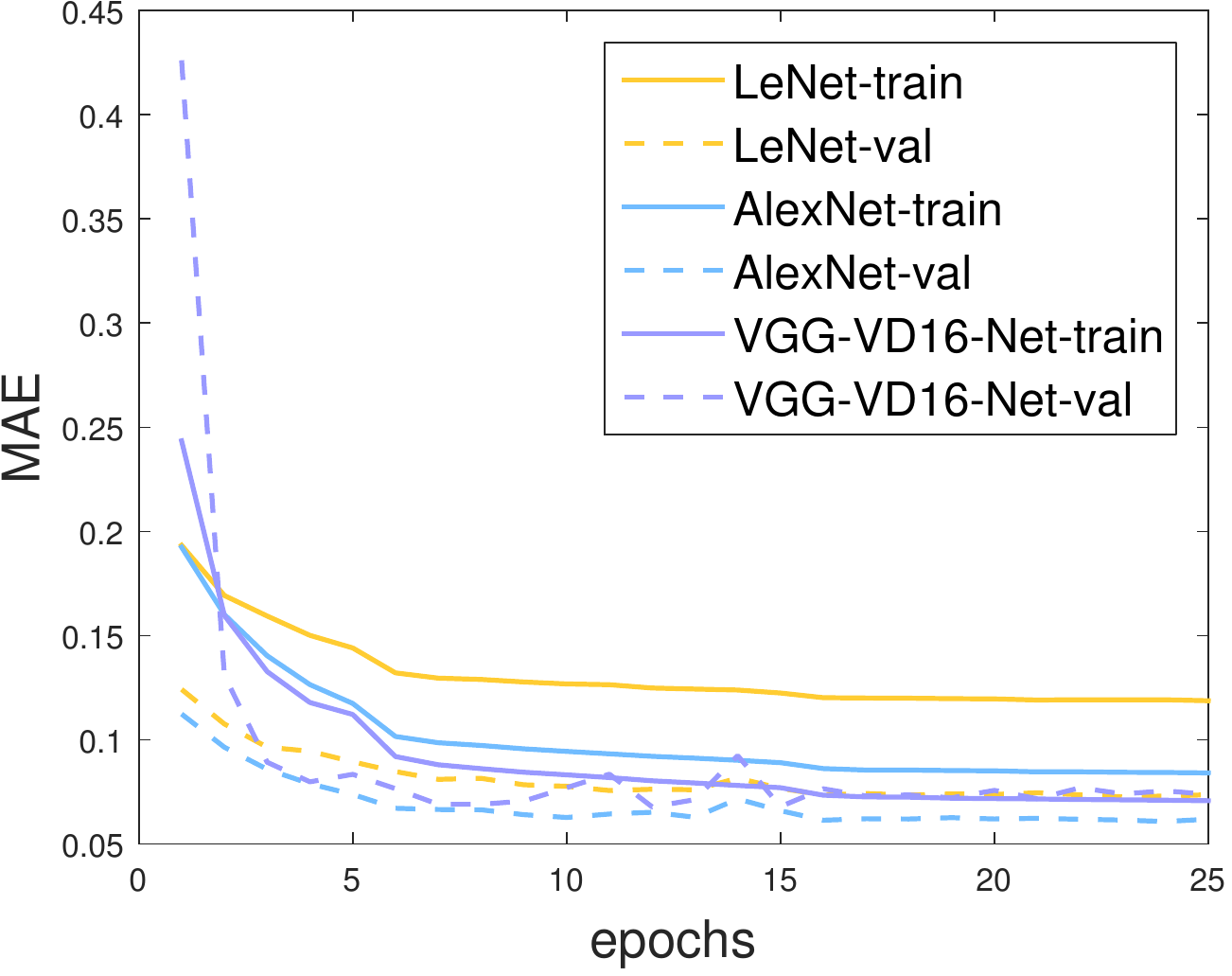}
	\caption{Training (train) and validation (val) errors in terms of $MAE$ versus the number of epochs on LeNet-like, AlexNet-like and VGG-VD16-like Tasselnet architectures.}
	\label{fig:learning_curves}
\end{figure}

\paragraph{Network architecture.} We first evaluate how the model capacity influence the results. As aforementioned, three network architectures of LeNet-like, AlexNet-like and VGG-VD16-Net-like Tasselnet are considered. The learning curves of three models are shown in Fig.~\ref{fig:learning_curves}. It is clear that, the deeper the model uses, the lower the training error achieves. However, a lower training error does not imply a lower validation error for our problem---the validation error of VGG-VD16-Net-like model shows obvious fluctuations and is also higher than the AlexNet-like model. Numerical results on the test set are shown in Table~\ref{tab:architecture}. Experimental results demonstrate that the AlexNet-like architecture achieves the overall best counting performance, with a $MAE$ of $6.6$ and a $MSE$ of $9.6$. The inferiority of LeNet and VGG-VD16-Net may boil down to the low model capacity and the over-fitting on the training data. This can be clearly observed in Fig.~\ref{fig:learning_curves}. The low model capacity of LeNet-like architecture shows a relatively high training error, which implies the data may be in the state of under-fitting. The VGG-VD16-Net-like architecture fits the training data well while exhibits a higher validation error (compared to the AlexNet-like architecture), which suggests the model may not generalize well on the test data. Based on these results, a moderately complex model seems sufficient for maize tassels counting.

\paragraph{Number of training samples.} Here we investigate how the number of training samples affects the counting performance. We vary the sampling stride $s_r$ using the range of values $2^n, n=5, 4, 3, 2$, leading to $2.37\times10^4$, $9.13\times10^4$, $3.56\times10^5$, and $1.41\times10^6$ training sub-images, respectively. Experimental results are listed in Table~\ref{tab:num_subimages}. We observe that the number of training samples indeed plays a vital role: $MAE$ decreases from $9.5$ to $6.5$ with increased training number of sub-images. In addition, the overall performance between $s_r=8$ and $s_r=4$ is almost identical, implying that a moderate number of training sub-images can already capture well the in-field variations.

\begin{table*}[!t] \footnotesize
	\centering
	\caption{Comparison of different loss functions for maize tassels counting on the MTC dataset. The lowest error is boldfaced.}
	\label{tab:loss}
	\renewcommand\arraystretch{1.25}
	\addtolength{\tabcolsep}{-3.5pt}
	\begin{tabular}{l|cc|cc|cc|cc|cc|cc|cc|cc|cc}
		\hline
		Loss                & \multicolumn{16}{c|}{Sequences} & \multicolumn{2}{c}{Overall}\\
		\hline
		& \multicolumn{2}{c|}{\scriptsize Zhengzhou2011}  & \multicolumn{2}{c|}{\scriptsize Taian2010\_2} & \multicolumn{2}{c|}{\scriptsize Taian2011\_2} & \multicolumn{2}{c|}{ \scriptsize Taian2012\_2} & \multicolumn{2}{c|}{\scriptsize Taian2013\_2} & \multicolumn{2}{c|}{\scriptsize Gucheng2014} & \multicolumn{2}{c|}{\scriptsize Jalaid2015\_2} & \multicolumn{2}{c|}{\scriptsize Jalaid2015\_3} & \\
		& MAE  & MSE  & MAE  & MSE  & MAE  & MSE  & MAE  & MSE  & MAE  & MSE  & MAE  & MSE  & MAE  & MSE  & MAE  & MSE  & MAE  & MSE \\
		Huber($\delta=0.1$) & 5.2 & 6.2 & 7.1 & 8.2 & 4.8 & 5.4 & 5.9 & 7.2 & 7.6 & 8.7 & 7.2 & 8.4 & 14.2 & 15.4 & 29.7 & 34.6 & 8.5 & 12.2\\
		Huber($\delta=1$)   & \textbf{3.9} & 4.9 & 4.1 & 5.1 & 4.7 & 5.5 & 3.8 & 4.4 & 8.7 & 11.1 & 9.6 & 11.1 & 11.1 & 12.2 & 22.3 & 26.5 & 7.5 & 10.5\\
		Huber($\delta=10$)  & 4.2 & 5.1 & 4.4 & 5.5 & 4.7 & 5.4 & 3.5 & 4.2 & 9.0 & 11.4 & 10.0 & 11.5 & \textbf{9.6} & \textbf{11.0} & \textbf{18.5} & \textbf{23.3} & 7.3 & 10.0\\
		$\ell_2$  & 4.0 & \textbf{4.8} & \textbf{3.9} & \textbf{4.9} & 4.9 & 5.6 & \textbf{3.3} & \textbf{3.9} & 7.7 & 9.4 & 9.4 & 10.8 & 10.3 & 11.4 & 23.5 & 27.3 & 7.3 & 10.3\\
		$\ell_1$  & 4.9 & 6.1 & 5.2 & 6.6 & \textbf{2.5} & \textbf{2.9} & 4.8 & 5.8 & \textbf{4.0} & \textbf{5.0} & \textbf{5.3} & \textbf{6.5} & 16.0 & 16.6 & 20.7 & 25.2 & \textbf{6.6} & \textbf{9.6}\\
		\hline
	\end{tabular}
\end{table*}

\begin{table*}[!t] \footnotesize
	\centering
	\caption{Comparison of different Gaussian kernel parameter $\sigma$ for maize tassels counting on the MTC dataset. The lowest error is boldfaced.}
	\label{tab:sigma}
	\renewcommand\arraystretch{1.25}
	\addtolength{\tabcolsep}{-3.5pt}
	\begin{tabular}{l|cc|cc|cc|cc|cc|cc|cc|cc|cc}
		\hline
		$\sigma$                & \multicolumn{16}{c|}{Sequences} & \multicolumn{2}{c}{Overall}\\
		\hline
		& \multicolumn{2}{c|}{\scriptsize Zhengzhou2011}  & \multicolumn{2}{c|}{\scriptsize Taian2010\_2} & \multicolumn{2}{c|}{\scriptsize Taian2011\_2} & \multicolumn{2}{c|}{ \scriptsize Taian2012\_2} & \multicolumn{2}{c|}{\scriptsize Taian2013\_2} & \multicolumn{2}{c|}{\scriptsize Gucheng2014} & \multicolumn{2}{c|}{\scriptsize Jalaid2015\_2} & \multicolumn{2}{c|}{\scriptsize Jalaid2015\_3} & \\
		& MAE  & MSE  & MAE  & MSE  & MAE  & MSE  & MAE  & MSE  & MAE  & MSE  & MAE  & MSE  & MAE  & MSE  & MAE  & MSE  & MAE  & MSE \\
		$\sigma=4$ & 5.2 & 6.5 & \textbf{4.9} & \textbf{6.4} & \textbf{2.3} & 3.3 & 5.5 & 6.4 & 4.5 & 6.0 & 4.2 & 5.4 & 18.6 & 19.5 & 27.4 & 32.6 & 7.0 & 11.3\\
		$\sigma=8$ & 4.9 & 6.1 & 5.2 & 6.6 & 2.5 & \textbf{2.9} & \textbf{4.8} & \textbf{5.8} & \textbf{4.0} & \textbf{5.0} & 5.3 & 6.5 & 16.0 & 16.6 & \textbf{20.7} & \textbf{25.2} & \textbf{6.6} & \textbf{9.6}\\
		$\sigma=12$ & \textbf{4.6} & \textbf{5.5} & 7.9 & 8.7 & 4.0 & 4.7 & 6.6 & 7.9 & 5.9 & 6.9 & \textbf{4.0} & \textbf{4.7} & \textbf{15.2} & \textbf{15.9} & 27.1 & 30.9 & 7.6 & 10.9\\
		\hline
	\end{tabular}
\end{table*}

\begin{table*}[!t] \footnotesize
	\centering
	\caption{Comparison of different sub-image sizes for maize tassels counting on the MTC dataset. The lowest error is boldfaced.}
	\label{tab:subim_sizes}
	\renewcommand\arraystretch{1.25}
	\addtolength{\tabcolsep}{-3.5pt}
	\begin{tabular}{l|cc|cc|cc|cc|cc|cc|cc|cc|cc}
		\hline
		$r\times r$                & \multicolumn{16}{c|}{Sequences} & \multicolumn{2}{c}{Overall}\\
		\hline
		& \multicolumn{2}{c|}{\scriptsize Zhengzhou2011}  & \multicolumn{2}{c|}{\scriptsize Taian2010\_2} & \multicolumn{2}{c|}{\scriptsize Taian2011\_2} & \multicolumn{2}{c|}{ \scriptsize Taian2012\_2} & \multicolumn{2}{c|}{\scriptsize Taian2013\_2} & \multicolumn{2}{c|}{\scriptsize Gucheng2014} & \multicolumn{2}{c|}{\scriptsize Jalaid2015\_2} & \multicolumn{2}{c|}{\scriptsize Jalaid2015\_3} & \\
		& MAE  & MSE  & MAE  & MSE  & MAE  & MSE  & MAE  & MSE  & MAE  & MSE  & MAE  & MSE  & MAE  & MSE  & MAE  & MSE  & MAE  & MSE \\
		$16\times 16$ & 4.6 & 5.6 & 10.1 & 11.4 & 10.1 & 13.0 & 7.5 & 11.5 & 6.1 & 7.5 & 8.2 & 10.1 & \textbf{15.7} & \textbf{16.5} & 26.9 & 31.7 & 9.9 & 13.4\\

		$32\times 32$ & 4.9 & 6.1 & 5.2 & 6.6 & \textbf{2.5} & \textbf{2.9} & 4.8 & 5.8 & \textbf{4.0} & \textbf{5.0} & 5.3 & 6.5 & 16.0 & 16.6 & \textbf{20.7} & \textbf{25.2} & \textbf{6.6} & \textbf{9.6}\\
		$64\times 64$ & 4.7 & 5.7 & 5.5 & 6.5 & 3.8 & 4.5 & 3.4 & 4.0 & 5.4 & 6.3 & \textbf{4.2} & \textbf{5.4} & 16.7 & 17.6 & 27.3 & 31.9 & 6.8 & 10.8\\

		$96\times 96$ & \textbf{4.4} & \textbf{5.2} & \textbf{4.2} & \textbf{5.3} & 3.6 & 4.2 & \textbf{2.9} & \textbf{3.5} & 7.5 & 10.1 & 4.9 & 5.8 & 15.9 & 17.7 & 30.5 & 35.2 & 6.9 & 11.5\\

		\hline
	\end{tabular}
\end{table*}

\paragraph{Loss function.} Here we compare the effect of different loss functions. As aforementioned, $\ell_1$ loss, $\ell_2$ loss, and $Huber$ loss are evaluated. $Huber$ loss contains a free parameter $\delta$, so we further add three variants of $Huber$ loss when $\delta=0.1$, $\delta=1$, and $\delta=10$. The lower $\delta$ is, the more $Huber$ loss looks like $\ell_1$ loss. The higher $\delta$ is, the more it looks like $\ell_2$ loss. Results are shown in Table~\ref{tab:loss}. We observe that, there is no single loss can achieve consistently better results than other competitors over all test sequences. This may have something to do with the problem nature of maize tassels counting and specific data distributions of test sequences. $Huber$ loss with $delta=10$ shows better performance on the two challenging sequences, which suggests that $Huber$ loss is indeed robust to noise but at a cost of sacrificing the ability to fit normal data samples (poor performance on the Taian2013\_2 and Gucheng2014 sequences). $Huber$ loss also has a problem that there is no principal way to choose an appropriate $\delta$ (the performance degrades when $delta=0.1$). The counting performance of $\ell_1$ loss and $\ell_2$ loss is comparable, but $\ell_1$ is generally more stable.

\paragraph{Gaussian kernel parameter.} The sensitivity of Gaussian kernel parameter $\sigma$ is further evaluated. Concretely, we set $\sigma=4$, $\sigma=8$, and $\sigma=12$, respectively. The results are listed in Table~\ref{tab:sigma}. We observe that the optimal $\sigma$ for each test sequence is different. The reason perhaps is that a fixed $\sigma$ cannot describe appropriately maize tassels of different sizes (due to different cultivars). Although the optimal counting performance cannot be achieved with a specific $\sigma$, the counting performance with different $\sigma$ does not vary significantly, which suggests that the mechanism of local counts regression is not that sensitive to specific choices of $\sigma$. Empirically, one can set an appropriate $\sigma$ by observing the Gaussian smoothing responses on the training set. The responses should fit the median size of maize tassels.

\paragraph{Sub-image sizes.} The influence of different sub-image sizes is also analysed. We compare the performance of four settings, including $r=16$, $r=32$, $r=64$, and $r=96$. Table~\ref{tab:subim_sizes} lists the results. According to the results, we again observe that the optimal performance for each test sequence does not correlate well with sub-image sizes. We think this has something to do with specific tassel sizes in each sequences. In practice, drawing upon a relatively small (but not too small) sub-image sizes is preferable. This is not just because one can densely sample a sufficient number of training samples but also because the variations within a small receptive field are easily modelled.

\begin{figure*}[h!]
	\centering
	\includegraphics[width=0.98\textwidth]{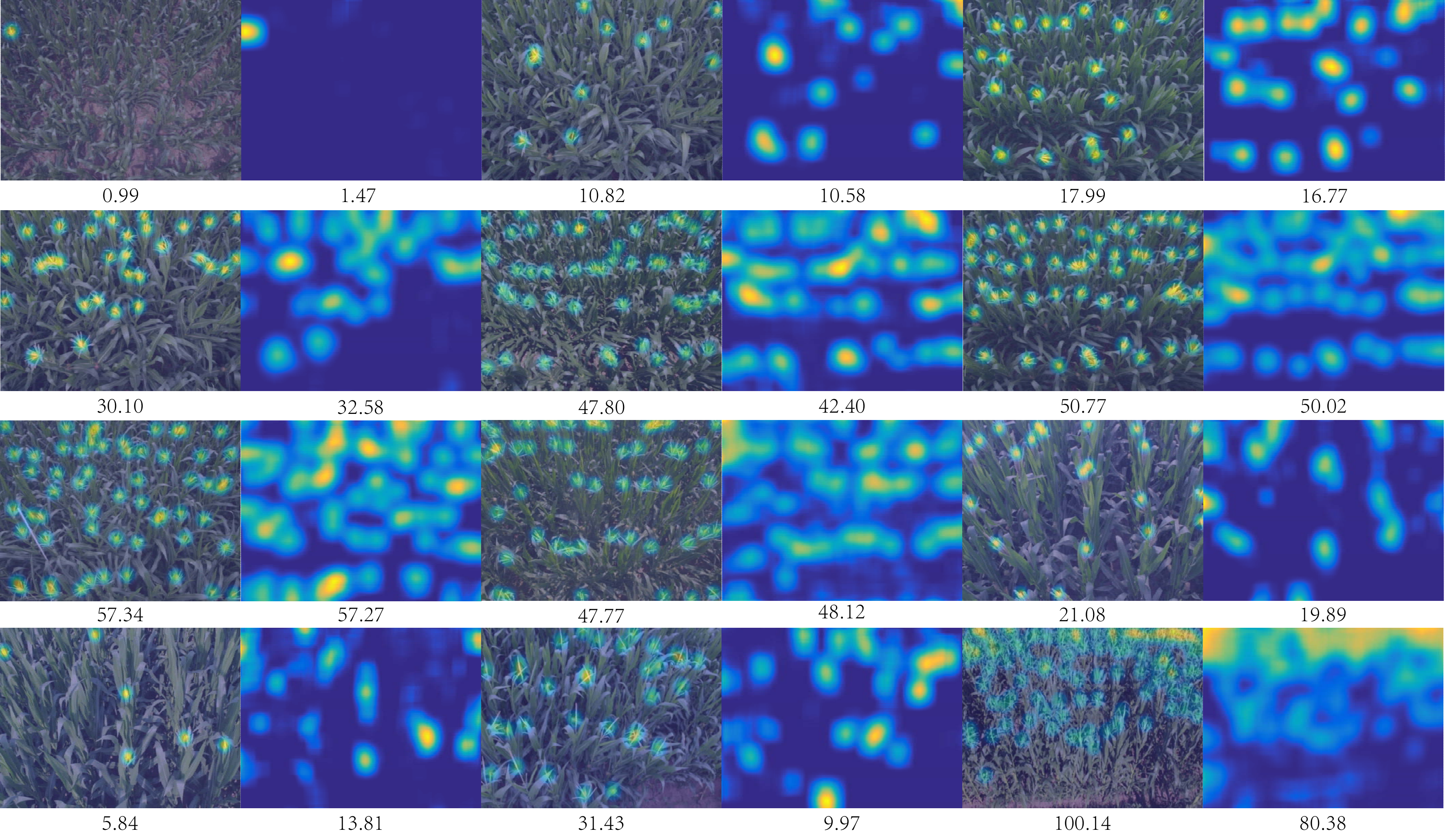}
	\caption{Qualitative results of ground truth density maps overlaid on original images and counting maps predicted by Tasselnet. The number shown below each sub-figure denotes the tassel count integrated over the density/count map. The last row shows three unsuccessful predictions.}
	\label{fig:count_maps}
\end{figure*}

\begin{table*}[h!] \footnotesize
	\centering
	\caption{Mean absolute errors (MAE) and mean squared errors (MSE) for maize tassels counting on the MTC dataset. The lowest error is boldfaced.}
	\label{tab:results}
	\renewcommand\arraystretch{1.25}
	\addtolength{\tabcolsep}{-3.5pt}
	\begin{tabular}{l|cc|cc|cc|cc|cc|cc|cc|cc|cc}
		\hline
		Method                & \multicolumn{16}{c|}{Sequences} & \multicolumn{2}{c}{Overall}\\
		\hline
		& \multicolumn{2}{c|}{\scriptsize Zhengzhou2011}  & \multicolumn{2}{c|}{\scriptsize Taian2010\_2} & \multicolumn{2}{c|}{\scriptsize Taian2011\_2} & \multicolumn{2}{c|}{ \scriptsize Taian2012\_2} & \multicolumn{2}{c|}{\scriptsize Taian2013\_2} & \multicolumn{2}{c|}{\scriptsize Gucheng2014} & \multicolumn{2}{c|}{\scriptsize Jalaid2015\_2} & \multicolumn{2}{c|}{\scriptsize Jalaid2015\_3} & \\
		& MAE  & MSE  & MAE  & MSE  & MAE  & MSE  & MAE  & MSE  & MAE  & MSE  & MAE  & MSE  & MAE  & MSE  & MAE  & MSE  & MAE  & MSE \\
		JointSeg	          & 20.9 & 23.2 & 46.6 & 47.9 & 16.4 & 19.7 & 25.1 & 29.8 & 6.5  & 8.0  & 7.3 & 10.5 & 27.8  & 29.1 & 53.2 & 61.3 & 24.2 & 31.6\\
		mTASSEL               & 9.8  & 14.9 & 18.6 & 22.1 & 11.6 & 12.7 & 5.3  & 7.8  & 13.1 & 16.6 & 31.1 & 35.3 & 16.2 & 18.0 & 46.6 & 51.0 & 19.6 & 26.1\\
		GlobalReg             & 19.0 & 21.5 & 23.0 & 24.7 & 14.1 & 16.8 & 13.5 & 15.7 & 19.6 & 25.2 & 19.5 & 21.7 & 11.2 & 13.7 & 42.1 & 45.4 & 19.7 & 23.3\\
		DensityReg            & 16.1 & 20.2 & 9.9  & 10.7 & 9.2  & 11.7 & 10.8 & 12.7 & 20.2 & 23.7 & 9.4  & 10.5 & \textbf{7.2}  & \textbf{7.9}  & 23.5 & 26.9 & 11.9 & 14.8\\
		CCNN                  & 21.3 & 23.3 & 28.9 & 31.6 & 12.4 & 16.0 & 12.6 & 15.3 & 18.9 & 23.7 & 21.6 & 24.1 & 9.6 & 12.4 & 39.5 & 46.4 & 21.0 & 25.5\\

		Tasselnet		      & \textbf{4.9} & \textbf{6.1} & \textbf{5.2} & \textbf{6.6} & \textbf{2.5} & \textbf{2.9} & \textbf{4.8} & \textbf{5.8} & \textbf{4.0} & \textbf{5.0} & \textbf{5.3} & \textbf{6.5} & 16.0 & 16.6 & \textbf{20.7} & \textbf{25.2} & \textbf{6.6} & \textbf{9.6}\\
		\hline
	\end{tabular}
\end{table*}

\subsection*{Comparison with the state of the art}
To place Tasselnet in the context of the state of the art, several well-established baseline approaches are chosen for comparison, they are:
\begin{itemize}
	\item \textit{JointSeg}~\cite{lu2016region}: JointSeg is the state-of-the-art tassel segmentation method. The number of object counts can be easily inferred from the segmentation results. We further perform some morphological operations as post-correction to reduce the segmentation noises. This approach can be viewed as a counting-by-segmentation baseline. It is not specially designed for a counting problem, but the comparison somewhat justify whether our problem could be addressed by a simple image processing technique.
	\item \textit{mTASSEL}~\cite{lu2015fine}: mTASSEL is the state-of-the-art tassel detection approach designed specifically for maize tassels. mTASSEL uses multi-view representations to characterise the visual characteristics of tassels to achieve robust detection. This is a counting-by-detection baseline.
	\item \textit{GlobalReg}~\cite{tota2015counting}: GlobalReg directly regresses the global count of images. Off-the-shelf fully-connected deep activations extracted from a pre-trained model are used as a holistic image representation. Then the global image feature is linearly mapped into a global object count by ridge regression. This is a global counting-by-regression baseline.
	\item \textit{DensityReg}~\cite{vlaz2010denlearn}: DensityReg is the seminal work that proposes the idea of density map regression. It predicts a count density for every pixel by optimising a so-called MESA distance. This is a global density-based counting-by-regression baseline.
	\item \textit{Counting-CNN (CCNN)}~\cite{onoro2016towards}: CCNN is a state-of-the-art object counting approach. It treats the local density map as the regression target and also uses a AlexNet-like CNN architecture. This is a local density-based counting-by-regression baseline.
\end{itemize}
Qualitative and quantitative results are shown in Fig.~\ref{fig:count_maps} and Table~\ref{tab:results}, respectively. Results of Tasselnet are reported using the default parameters setting. According to the results, we can make the following observations:
\begin{itemize}
	\item Tasselnet outperforms other baseline approaches in $7$ out of $8$ test sequences and achieves the overall best counting performance---$MAE$ and $MSE$ are significantly lower than other competitors.
	\item The poor performance of JointSeg and mTASSEL implies that the problem of in-field counting of maize tassels cannot be solved by simple colour-cue-based image segmentation or standard object detection.
	\item Even a simple global regression can achieve comparable counting performance against mTASSEL in which the bounding-box-level annotations are utilized. This suggests it is better to formulate the problem of maize tassels counting in a counting-by-regression manner.
	\item Regressing the global density map can also reduce the counting error effectively. However, it is hard to extend this idea to the deep CNN-based paradigm, because there is currently no dataset with thousands of labelled images samples to make the learning of deep networks tractable, especially in the plant-related scenarios. Hence, DensityReg cannot enjoy the bonus brought by deep CNN, and the performance may be limited by the power of feature representation.
	\item The performance of CCNN even falls behind the global regression baseline. In experiments we observe that CCNN performs poorly when given an image with just a few tassels of different types. Compared to regressing local counts as in Tasselnet, CCNN needs to fit harsher pixel-level ground truth density, so it likely suffers in the vague definition of density map due to different tassel sizes. This may explain why local density regression does not work when given varying object sizes like maize tassels.
	\item Qualitative results in Fig.~\ref{fig:count_maps} show that Tasselnet can give reasonable approximations to the ground truth density maps. In most cases, the estimated counts are similar to the ground truth counts. However, there also exists some circumstances that Tasselnet cannot give an accurate prediction. The last row in Fig.~\ref{fig:count_maps} shows three failure cases: i) When the image is captured under extremely strong illuminations, highlight regions of leaves will contribute to several fake responses; ii) If maize tassels present long-tailed shapes in images, the long-tailed parts only receive partial local counts, resulting in a under-estimate situation; iii) The extremely crowded scene is also beyond the ability of Tasselnet. To alleviate these issues, one may consider to add extra training data that contain the extremely crowded scenarios. Alternatively, since the training sequences and test sequences exhibit more or less different data distributions, it may be possible to use domain adaptation~\cite{Lu2017} to fill the last few percent of difference between sequences. We leave these as the future explorations of this work.
\end{itemize}

As a summary of our evaluations, we suggest the following good practices for maize-tassel-like in-field counting problems:
\begin{enumerate}
	\item Try the idea of counting by regression if the objects exhibit significant occlusions.
	\item Try local counts regression if the physical size of objects varies dramatically.
	\item Use a relatively small sub-image size so that a sufficient number of training samples could be sampled.
	\item It is safe to use a moderately complex deep model.
	\item Try $\ell_1$ loss first to achieve a robust regression.
\end{enumerate}

\section*{Conclusions}
In this paper, we rethink the problem nature of in-field counting of maize tassels and novelly formulates the problem as an object counting task. A tailored MTC dataset with 361 field images captured during 6 years and corresponding manually-labelled dotted annotations is constructed. An effective deep CNN-based solution, Tasselnet, is also presented to count effectively maize tassels via local counts regression. We show that local counts regression is particularly suitable for counting problems whose ground truth density maps cannot be precisely defined. Extensive experiments are conducted to justify the effectiveness of our proposition. Results show that Tasselnet achieves the state-of-the-art performance and outperforms previous baseline approaches by large margins.

For future work, we will continue to enrich the MTC dataset, because the training data are always the key to the good performance, especially the data diversity. In addition, we will explore the feasibility to improve the counting performance in the context of domain adaptation, because the adaptation of object counting problems still remains an open question. In-field counting of maize tassels is a challenging problem, not only because the unconstrained natural environment but also because the self-changing rule of plants growth. We hope this paper could attract interests of both Plant Science and Computer Vision communities and inspires further studies to advance our knowledge and understanding towards the problem.

\begin{backmatter}

\section*{Declarations}

\section*{Ethics approval and consent to participate}
Not applicable

\section*{Consent for publication}
Not applicable

\section*{Competing interests}
The authors declare that they have no competing interests.

\section*{Availability of data and materials}
The MTC dataset and other supporting materials will be made available online.

\section*{Funding}
This work was supported in part by the Special Scientific Research Fund of Meteorological Public Welfare Profession of China under Grant GYHY200906033 and in part by the National Natural Science Foundation of China under Grant 61502187.

\section*{Author's contributions}
HL proposed the idea of counting maize tassels via local counts regression, implemented the technical pipeline, conducted the experiments, analysed the results, and drafted the manuscript. ZG, YX and CS co-supervised the study and contributed in writing the manuscript.
BZ helped to design the experiments and provided technical support
 for efficient model training.

 All authors read and approved the final manuscript.

\section*{Acknowledgements}
The authors would like to thank Xiu-Shen Wei for helpful discussions.
\bibliographystyle{unsrt} %
\bibliography{bmc_article}      %

\end{backmatter}
\end{document}